\documentclass{article}

\usepackage[nonatbib,preprint]{neurips_2019}
\usepackage[numbers,sort,compress]{natbib}

\usepackage[utf8]{inputenc} 
\usepackage[T1]{fontenc}    
\usepackage{url}            
\usepackage{booktabs}       
\usepackage{amsfonts}       
\usepackage{nicefrac}       
\usepackage{microtype}      

\usepackage{math}
\usepackage{enumitem}

\usepackage{xspace, amsmath, amssymb, hyperref,subcaption}
\usepackage{bm}
\usepackage{graphicx}
\usepackage{color}
\usepackage{url}

\usepackage{mathtools}

\newcommand{\bx}{\mathbf{x}}
\newcommand{\bs}{\mathbf{s}}

\newcommand{\ba}{\mathbf{a}}

\newcommand{\bz}{\mathbf{z}}

\newcommand{\M}{\mathcal{M}}
\newcommand{\bi}[1]{\textit{\textbf{#1}}}

\newcommand{\gnef}{g_\textup{nef}}
\newcommand{\gper}{g_\textup{per}}

\title{Online Data Poisoning Attacks}

\author{
		Xuezhou Zhang\\
		Department of Computer Sciences\\
		University of Wisconsin--Madison\\
		\texttt{zhangxz1123@cs.wisc.edu}
		\And
		Xiaojin Zhu\\
		Department of Computer Sciences\\
		University of Wisconsin--Madison\\
		\texttt{jerryzhu@cs.wisc.edu}  
		\And
		Laurent Lessard\\
		Department of Electrical and Computer Engineering\\
		University of Wisconsin--Madison\\
		\texttt{laurent.lessard@wisc.edu}
}

\begin{document}

\maketitle

\begin{abstract}
	We study data poisoning attacks in the online setting where training items arrive sequentially,
and the attacker may perturb the current item to manipulate online learning. Importantly, the attacker has no knowledge of future training items nor the data generating distribution. 
	We formulate online data poisoning attack as a stochastic optimal control problem, and solve it with model predictive control and deep reinforcement learning.
We also upper bound the suboptimality suffered by the attacker for not knowing the data generating distribution. 
	Experiments validate our control approach in generating near-optimal attacks on both supervised and unsupervised learning tasks.
\end{abstract}

\section{Problem Statement}
\label{sec:problem}
Protecting machine learning from adversarial attacks is of paramount importance~\cite{vorobeychik2018adversarial,joseph2018adversarial,zhu2018optimal}. To do so one much first understand various types of adversarial attacks. 
Data poisoning is a type of attack where an attacker contaminates the training data in order to force a nefarious model on the learner~\cite{xiao2015support,mei2015using,burkard2017analysis,chen2019optimal,jun2018adversarial,li2016data}. 
Prior work on data poisoning focused almost exclusively on the batch setting, where the attacker poisons a batch training set, and then the victim learns from the batch~\cite{biggio2012poisoning,munoz2017towards,xiao2015support,mei2015using,sen2018training,chen2017targeted}. 
However, the batch setting misses the threats posed by the attacker on sequential learners.
For example, in e-commerce applications user-generated data arrives at the learner sequentially. Such applications are particularly susceptible to poisoning attacks, because it is relatively easy for the attacker to manipulate data items before they arrive at the learner.
Furthermore, the attacker may observe the effect of previous poisoning on the learner and \emph{adaptively} decides how to poison next.
This adaptivity makes online data poisoning a potentially more severe threat compared to its batch counterpart.

This paper presents a principled study of online data poisoning attacks.
Our key contribution is an optimal control formulation of such attacks.
We provide theoretical analysis to show that the attacker can attack near-optimally even without full knowledge of the underlying data generating distribution. We then propose two practical attack algorithms---one based on traditional model-based optimal control, and the other based on deep reinforcement learning---and show that they achieve near-optimal attack performance in synthetic and real-data experiments.
Taken together, this paper builds a foundation for future studies of defense against online data poisoning. 

\begin{figure}[ht]
	\centering
	\includegraphics{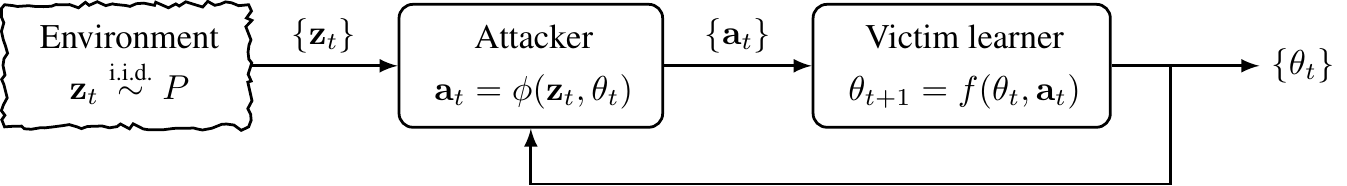}
	\caption{Online data poisoning attack diagram. The attacker observes the training samples $\{\bz_t\}$ and the learner's model $\{\theta_t\}$ in an online fashion, and injects poisoned samples $\{\ba_t\}$.}
	\label{fig:threatmodel}
\end{figure}
The online data poisoning problem in this paper is shown in Figure~\ref{fig:threatmodel}.
There are three entities: a stochastic \bi{environment}, a sequential learning \bi{victim}, and the online \bi{attacker}. 
In the absence of attacks,	
at time $t$ the environment draws a training data point $\bz_t \in \mathcal Z$ i.i.d.\ from a time-invariant distribution $P$: 
$\bz_t \stackrel{\text{i.i.d.}}{\sim} P.$
For example, $\bz_t$ can be a feature-label pair $\bz_t\defeq(\bx_t,y_t)$ in supervised learning or just the features $\bz_t\defeq\bx_t$ in unsupervised learning.  
The victim maintains a model $\theta_t\in\Theta$. 
Upon receiving $\bz_t$, the victim performs one step of the sequential update defined by the function $f$:
\begin{equation}
\theta_{t+1} = f(\theta_t,\bz_t), 
\label{eq:f}
\end{equation}
For example, $f$ can be gradient descent $f(\theta_t,\bz_t) \defeq \theta_t - \eta \nabla \ell(\theta_t, \bz_t)$ under learner loss $\ell$ and step size $\eta$.
We now introduce the attacker by defining its knowledge, allowed actions, and goals:
\begin{itemize}[leftmargin=*]
	
	\item The attacker has knowledge of
	the victim's update function $f$, the victim's initial model $\theta_0$, data $\bz_{0:t}$ generated by the environment so far, and optionally $n$ ``pre-attack'' data points $\bz_{-n:-1} \stackrel{\text{i.i.d.}}{\sim} P$. 
	\textbf{However, at time $t$ the attacker does not have the clairvoyant knowledge of future data points} $\bz_{t+1}, \bz_{t+2}, \ldots$, nor does it have the knowledge of the environment distribution $P$.

	\item The attacker can perform only one type of action:
	once the environment draws a data point $\bz_t$, the attacker perturbs the data point into a potentially different point $\ba_t \in \mathcal Z$.  
	The attacker incurs a {perturbation cost} $\gper(\bz_t,\ba_t)$, which reflects the price to attack.
	For example, $\gper(\bz_t,\ba_t) \defeq \|\ba_t-\bz_t\|_p$ if $\mathcal Z$ is endowed with an appropriate $p$-norm. 
	The attacker then gives $\ba_t$ to the victim, who proceeds with model update~\eqref{eq:f} using $\ba_t$ instead of $\bz_t$.

	\item The attacker's goal, informally, is to force the victim's learned models $\theta_t$ to satisfy certain nefarious properties at each step while paying a small cumulative perturbation cost.
	These ``nefarious properties'' (rather the inability to achieve them) are captured by a nefarious cost $\gnef(\theta)$.  It can encode a variety of attack goals considered in the literature such as: 
	(\textit{i}) \textit{targeted attack} $\gnef(\theta) \defeq \|\theta - \theta^{\dagger}\|$ to drive the learned model toward an attacker-defined target model $\theta^{\dagger}$ (the dagger is a mnemonic for attack); 
	(\textit{ii}) \textit{aversion attack} $\gnef(\theta) \defeq -\|\theta - \hat\theta\|$ (note the sign) to push the learned model away from a good model $\hat\theta$, such as the one estimated from pre-attack data;
	(\textit{iii}) \textit{backdoor attack} $\gnef(\theta) \defeq \ell(\theta, \bz^\dagger)$, in which the goal is to plant a backdoor such that the learned model behaves unexpectedly on special examples $\bz^\dagger$~\cite{li2016data,sen2018training,chen2017targeted}.
	To balance nefarious properties with perturbation cost, the attacker defines a \emph{running cost} $g$ at time $t$:
	\begin{equation}
	g(\theta_{t},\bz_t,\ba_t)\defeq \lambda \gnef(\theta_{t}) + \gper(\bz_t,\ba_t),
	\label{eq:g_t}
	\end{equation}
	where $\lambda$ is a weight chosen by the attacker to balance the two.
	The attacker desires small cumulative running costs, which is the topic of Section~\ref{sec:optimalcontrol}.
\end{itemize}


\section{Related Work}
Data poisoning attacks have been studied against a wide range of learning systems. 
However, this body of prior work has almost exclusively focused on the offline or batch settings, where the attacker observes and can poison the whole training set or an entire batch of samples at once, respectively.
In contrast, our paper focuses on the online setting, where the attacker has to act multiple times and sequentially during training.
Examples of offline or batch poisoning attacks against SVM include~\cite{biggio2012poisoning,burkard2017analysis,xiao2015support}. 
Such attacks are generalized into a bilevel optimization framework against general offline learners with a convex objective function in~\cite{mei2015using}.
A variety of attacks against other learners have been developed, including neural networks \cite{koh2017understanding,munoz2017towards},
autoregressive models \cite{alfeld2016data,chen2019optimal},
linear and stochastic bandits \cite{jun2018adversarial,ma2018data},
collaborative filtering \cite{li2016data}, and models for
sentiment analysis \cite{newell2014practicality}. 

There is an intermediate attack setting between offline and online, which we call \textbf{clairvoyant online attacks}, where the attacker performs actions sequentially but \textbf{has full knowledge of all future input data} $\bz_{t+1}, \bz_{t+2}, \ldots$
Examples include heuristic attacks against SVM learning from data streams~\cite{burkard2017analysis} and binary classification with an online gradient descent learner~\cite{wang2018data}.
Our paper focuses instead on the perhaps more realistic setting where the attacker has no knowledge of the future data stream.
More broadly, our paper advocates for a general optimal control viewpoint that is not restricted to specific learners such as SVM.

%

The parallel line of work studying \textbf{online teaching} also considers the sequential control problem of machine learners, where the goal is to choose a sequence of training examples that accelerates learning~\cite{liu2017iterative,lessard2018optimal}. However, \cite{liu2017iterative} solves the problem using a greedy heuristic that we show in Section~\ref{sec:exp} performs poorly compared to our optimal control approach. On the other hand, \cite{lessard2018optimal} finds optimal teaching sequences but is restricted to an ordinary linear regression learner.

The problem of optimal feedback control in the presence of noise, uncertain disturbances, or uncertain dynamics has been an area of study for the better part of the past century. Major subfields include \textit{stochastic control}, when disturbances are stochastic in nature~\cite{aastrom2012introduction,kumar2015stochastic}, \textit{adaptive control}, when unknown parameters must be learned in an online fashion~\cite{aastrom2013adaptive,sastry1989adaptive}, and \textit{robust control}, when a single controller is designed to control a family of systems within some uncertainty set~\cite{zhou1996robust,skogestad2007multivariable}.

More recently, these classical problems have been revisited in the context of modern statistics, with the goal of obtaining tight sample complexity bounds. Examples include unknown dynamics \cite{dean2017sample}, adversarial dynamics \cite{agarwal2019online}, adversarial cost \cite{cohen2018online} or unknown dynamics \textit{and} cost \cite{fiechter1997pac}.
These works typically restrict their attention to linear systems with quadratic or convex losses, which is a common and often reasonable assumption for control systems.
However, for our problem of interest described in Section~\ref{sec:problem}, the system dynamics~\eqref{eq:f} are the learner's dynamics, which are nonlinear for all cases of practical interest, including simple cases like gradient descent. In the sections that follow, we develop tools, algorithms, and analysis for handling this more general nonlinear setting.

\section{An Optimal Control Formulation} 
\label{sec:optimalcontrol}
We now precisely define the notion of optimal online data poisoning attacks.
To do so, we cast the online data poisoning attack setting in Section~\ref{sec:problem} as a \bi{Markov Decision Process (MDP)} $\M = (S,A,T,g,\gamma,\bs_0)$ explained below.  
\begin{itemize}[leftmargin=*]
	\item
	The \textbf{state} $\bs_t$ at time $t$ is the stacked vector  $\bs_t \defeq [\theta_t, \bz_t]^\tp$ consisting of the victim's current model $\theta_t$ and the incoming environment data point $\bz_t$.
	The \textbf{state space} is $S \defeq \Theta \times \mathcal Z$.
	\item
	The attacker's \textbf{action} is the perturbed training point, i.e.\  $\ba_t \in \mathcal Z$.
	The \textbf{action space} is $A\defeq\mathcal Z$.
	\item
	From the attacker's perspective, the \textbf{state transition probability} $T:S\times A \to \Delta_S$, where $\Delta_S$ is the probability simplex over $S$, describes the conditional probability on the next state given current state and attack action.
	Specifically, $T(\bs_{t+1} \mid \bs_t, \ba_t) = T([\theta_{t+1},\bz_{t+1}]^\tp \mid \bs_t, \ba_t) = Pr(f(\theta_t, \ba_t) = \theta_{t+1}) \cdot P(\bz_{t+1})$.
	For concreteness, in this paper, we assume that the victim learning update $f$ is \textbf{deterministic}, and thus the stochasticity is solely in the $\bz_{t+1}$ component inside $\bs_{t+1}$, which has a marginal distribution $P$, i.e.\
	\begin{eqnarray}
	T\!\left([f(\theta_t, \ba_t),\bz_{t+1}]^\tp \suchthat \bs_t, \ba_t\right) = P(\bz_{t+1}).\label{eq:T}
	\end{eqnarray}

	\item
	The quality of control at time $t$ is specified by the \textbf{running cost} $g(\theta_{t},\bz_t,\ba_t)$ in~\eqref{eq:g_t}, to be minimized. From now on, we overload the notation and write the running cost equivalently as $g(\bs_t,\ba_t)$. Note that this is the opposite of the reward maximization setup commonly seen in reinforcement learning.
	
	\item
	We present online data poisoning attack with an infinite time horizon (the finite horizon case is similar but omitted due to space).  
	We introduce a \textbf{discounting factor} $\gamma \in (0,1)$ to define a discounted cumulative cost. 
	\item
	The \textbf{initial probability} $\mu_0: S\to \Delta_S$ is the probability distribution of the initial state $\bs_0$. In particular, we assume that the initial model $\theta_0$ is fixed while the first data point $\bz_0$ is sampled from $P$, i.e.\ $\mu_0(\theta_0,\bz_0) = P(\bz_0)$.
\end{itemize}

A \textbf{policy} is a function $\phi: S \to A$ that the attacker uses to choose the attack action 
$\ba_t \defeq \phi(\bs_t) =\phi([\theta_t, \bz_t]^\tp )$ 
based on the current victim model $\theta_t$ and the current environment input $\bz_t$.
The \textbf{value} $V_\M^\phi(\bs)$ of a state $\bs$ is the expected discounted cumulative cost starting at $\bs$ and following policy $\phi$:
\begin{equation}
V_\M^\phi(\bs) \defeq \E_\M \sum_{t=0}^\infty \gamma^{t} g(\bs_t, \phi(\bs_t))\biggr|_{\bs_0 = \bs}
\label{eq:V}
\end{equation}
where the expectation is over the transition probability $T$.
Overall, the attacker wants to perform optimal control over the MDP, that is, to find an \textbf{optimal control policy} $\phi^\star_\M$ that minimizes the expected value at the initial state. Define the attacker's objective as 
$
J_\M(\phi) \defeq \E_{\bs\sim\mu_0} V_\M^\phi(\bs)
$
, and the attacker's optimal attack policy as
$
\phi^\star_\M = \argmin_{\phi} J_\M(\phi).
$

Fortunately for the victim, the attacker cannot directly solve this optimal attack problem because it does not know the environment data distribution $P$ and thus cannot evaluate the expectation.
However, as we show next, the attacker can use model predictive control to approximately and incrementally solve for the optimal attack policy while it gathers information about $P$ as the attack happens.


\section{Practical Attack Algorithms via Model Predictive Control}
\label{sec:algs}

The key obstacle that prevents the attacker from obtaining an optimal attack is the unknown data distribution $P$.
However, the attacker can build an increasingly accurate \textbf{empirical distribution} $\hat P_t$ from $\bz_{0:t}$ and optionally the pre-attack data sampled from $P$. 
Specifically, at time $t$ with $\hat P_t$ in place of $P$ and with the model $\theta_t$ in place of $\theta_0$, the attacker can construct a surrogate MDP $\hat \M_t = (S,A,\hat T_t,g,\gamma,\hat\mu_t)$,
solve for the optimal policy $\phi^\star_{\hat\M_t} = \argmin_{\phi}J_{\hat{\M_t}}(\phi)$ on $\hat \M_t$,
and use $\phi^\star_{\hat\M_t}$ to perform a \textbf{one-step attack}: $ \ba_t = \phi^\star_{\hat \M_t}(\bs_t)$.

As time $t$ goes on, the attacker repeats the process of estimating $\phi^\star_{\hat\M_t}$ and applying the one-step attack $\phi^\star_{\hat\M_t}(\bs_t)$.
This repeated procedure of (re)-planning ahead but only executing one action is called \bi{Model Predictive Control (MPC)}~\cite{borrelli2017predictive,mayne2000constrained}, and is widely used across the automotive, aerospace, and petrochemical industries, to name a few.
At each time step $t$, MPC would plan a sequence of attacks using the surrogate model (in our case $\hat P_t$ instead of $P$), apply the first attack $\ba_t$, update $\hat P$, and repeat. This allows the controller to continually adapt without committing to an inaccurate model.

Next, we present two algorithms that practically solve the surrogate MDP, one based on model-based planning and the other based on model-free reinforcement learning.

\subsection{Algorithm NLP: Planning with Nonlinear Programming}
In the NLP algorithm, the attacker further approximates the surrogate objective as
$J_{\hat{\M_t}}(\phi)\approx\E_{\hat P_t} \left[\sum_{\tau=t}^{t+h-1} \gamma^{\tau-t}g(\bs_\tau,\phi(\bs_\tau))\right]
\approx  \sum_{\tau=t}^{t+h-1} \gamma^{\tau-t}g(\bs_\tau,\ba_\tau) \Bigr|_{\bz_{t:t+h-1}}$.
The first approximation truncates at $h$ steps after $t$, making it a finite-horizon control problem. 
The second approximation does two things: 
(i) It replaces the expectation by one sampled trajectory of the future input sequence, i.e.\ $\bz_{t:t+h-1}\sim\hat P_t$.
It is of course possible to use the average of multiple trajectories to better approximate the expectation, though empirically we found that one trajectory is sufficient. 
(ii) Instead of optimizing over a policy $\phi$, it locally searches for the action sequence $\ba_{t:t+h-1} \in \mathcal Z$.
The attacker now solves the following optimization problem at every time $t$:
\begin{eqnarray}
\min_{\ba_{t:t+h-1}}&& \sum_{\tau=t}^{t+h-1} \gamma^{\tau-t}g(\bs_\tau,\ba_\tau)\\
\mbox{s.t. } &&\bs_{\tau+1} = [f(\bs_\tau,\ba_\tau), \bz_\tau]^\tp, \forall \tau = t,...,t+h-1\nonumber\\
&&\bz_{t:t+h-1}\mbox{ and } \bs_t \mbox{ fixed and given.}\nonumber
\end{eqnarray}
Let $\ba^\star_{t:t+h-1}$ be a solution.
The NLP algorithm \emph{defines} $\phi^\star_{\hat M_t}(\bs_t) \defeq \ba^\star_t$, then moves on to $t+1$.
The resulting attack problem in general has a nonlinear objective stemming from $\gnef()$
and $\gper()$ in~\eqref{eq:g_t}, and nonconvex equality constraints stemming from the victim's learning rule $f()$ in~\eqref{eq:f}.
Nonetheless, the attacker can solve modest-sized problems using modern nonlinear programming solvers such as IPOPT \cite{wachter2006implementation}.


\subsection{Algorithm DDPG: Deep Deterministic Policy Gradient}
Instead of truncating and sampling to approximate the surrogate attack problem with a nonlinear program, one can directly solve for the optimal parametrized policy $\phi$ using reinforcement learning. In this paper, we utilize \textbf{deep deterministic policy gradient (DDPG)} \cite{lillicrap2015continuous} to handle a continuous action space. 
DDPG learns a deterministic policy with an actor-critic framework. Roughly speaking, it simultaneously learns an \textbf{actor network} $\mu(s)$ parametrized by $\theta^\mu$ and a \textbf{critic network} $Q(s,a)$ parametrized by $\theta^Q$. The actor network represents the currently learned policy while the critic network estimate the action-value function of the current policy, whose functional gradient guides the actor network to improve its policy. Specifically, the policy gradient can be written as:
$\nabla_{\theta^\mu}J = \E_{\bs\sim\rho^\mu} [\nabla_a Q(s, \mu(s)|\theta^Q)\nabla_{\theta^\mu}\mu(s|\theta^\mu)]$
in which the expectation is taken over $\rho^\mu$, the \textit{discounted state visitation distribution} for the current policy $\mu$. The critic network is updated using Temporal-Difference (TD) learning. We refer the reader to 
the original paper \cite{lillicrap2015continuous} for a more detailed discussion of this algorithm and other deep learning implementation details.

There are two advantages of this policy learning approach to the direct approach NLP. Firstly, it actually learns a policy which can \textbf{generalize} to more than one step of attack. Secondly, it is a \textbf{model-free} method and doesn't require knowledge of the analytical form of the system dynamic $f$, which is necessary for the direct approach. Therefore, DDPG also applies to the black-box attack setting, where the learner's dynamic $f$ is unknown.
To demonstrate the generalizability of the learned policy, in our experiments described later, we only allow the DDPG method to train once at the beginning of the attack on the surrogate MDP $\hat\M_0$ based on $(\theta_0,\bz_0)$ and the pre-attack data $\bz_{-n:-1}$. The learned policy $\phi_{\hat\M_0}$ is then applied to all later attack rounds without retraining.

\section{Theoretical Analysis}
The fundamental statistical limit to a realistic attacker is its lack of knowledge on the environment data distribution $P$.
An idealized attacker with knowledge of $P$ can find the {optimal control policy} $\phi^\star_\M$ that achieves the optimal attack objective $J_\M$.
In contrast, a realistic attacker only has an estimated $\hat P$, hence an estimated state transition $\hat T$, and ultimately an estimated MDP $\hat \M = (S,A,\hat T,g,\gamma,\hat\mu_0)$. 
The realistic attacker will find an optimal policy with respect to its estimated MDP $\hat \M$: $\phi^\star_{\hat\M} = \argmin_{\phi} J_{\hat{\M}}(\phi)$, but $\phi^\star_{\hat\M}$ is in general suboptimal with respect to the true MDP $\M$. 
We are interested in the \textbf{optimality gap} $V_\M^{\phi^\star_{\hat{\M}}}(s) - V_\M^{\phi^\star_\M}(s)$.
Note both are evaluated on the true MDP.

We present a theoretical analysis relating the optimality gap to the quality of estimated $\hat P$.
Our analysis is a natural extension to the {Simulation Lemma} in tabular reinforcement learning~\cite{kearns2002near} and that of~\cite{azar2017minimax}.
We assume that both $\mathcal{Z}$ and $\Theta$ are compact, and the running cost $g$ is continuous and thus bounded on its compact domain. WLOG, we assume $g \in [0,C_{\max}]$.
It is easy to see that then the range of value is bounded: $V \in [0, {C_{\max} \over 1-\gamma}]$ for both $\M, \hat M$, any policy, and any state.
Note the value function~\eqref{eq:V} satisfies the \textbf{Bellman equation}: 
$V_\M^\phi(\bs) = g(\bs,\phi(\bs)) + \gamma \E_{T(\bs' | \bs, \phi(\bs))} V_\M^\phi(\bs')$.

\begin{prop}\label{thm:sim_lemma}Consider two MDPs $\M, \hat \M$ that differ only in state transition, induced by $P$ and $\hat P$, respectively.
	Assume that $\|\hat P -P\|_1 \defeq \int_{\mathcal Z} \bigl|\hat P(\bz) -P(\bz)\bigr|\,\mathrm{d}\bz \leq \epsilon$. Let $\phi^\star_\M$ denote the optimal policy on $\M$ and $\phi^\star_{\hat \M}$ the optimal policy on $\hat{\M}$. Then, 
	$\sup_{s\in S} V_\M^{\phi^\star_{\hat{\M}}}(s) - V_\M^{\phi^\star_\M}(s)  \leq \frac{\gamma C_{\max}\epsilon }{(1-\gamma)^2}$.
\end{prop}

Proposition~\ref{thm:sim_lemma} implies that optimality gap is at most linear in $\epsilon \defeq \|\hat P -P\|_1$.
Classic Results on Kernel Density Estimation (KDE) suggest that the $L_1$ distance between $P$ and the kernel density estimator $\hat P_n$ based on $n$ samples converges to zero asymptotically in a rate of $O(n^{-s/d+2s})$ for some constant $s$ (e.g.\ Theorem 9 in \cite{holmstrom1992asymptotic}). 

In the experiment section below, the environment data stream is generated from a uniform distribution on a finite data set, in which case $P$ is a multinomial distribution. 
Under this special setting, we are able to provide a finite-sample bound of order $O(n^{-1/2})$ that matches with the best achievable asymptotic rate above, i.e.\ as $s \rightarrow \infty$.

\begin{thm}
	\label{thm:mn_convergence}Consider an MDP $\M$ induced by a multinomial distribution $P$ with support cardinality $N$, and a surrogate MDP $\hat \M$ induced by the empirical distribution $\hat P$ on $n$ i.i.d.\ samples, i.e.\ $\hat P(i) = \frac{1}{n}\sum_{j=1}^n I_{x_j=i}$. Denote $\phi^\star_\M$ the optimal policy on $\M$ and $\phi^\star_{\hat \M}$ the optimal policy on $\hat{\M}$. Then, with probability at least $1-\delta$, we have
	$\sup_{s\in S} V_\M^{\phi^\star_{\hat{\M}}}(s) - V_\M^{\phi^\star_\M}(s)  \leq \frac{2\gamma C_{\max}}{(1-\gamma)^2}\sqrt{\frac{1}{2n}\ln{2^{N+1}\over \delta}} = O(n^{-1/2})$.
\end{thm}

\section{Experiments}
\label{sec:exp}

In this section, we empirically evaluate our attack algorithms NLP and DDPG in Section~\ref{sec:algs} against several baselines on synthetic and real data.
As an empirical measure of attack efficacy, we compare the  attack methods by their \textbf{empirical discounted cumulative cost} $\tilde J(t) \defeq \sum_{\tau=0}^t \gamma^\tau g(\theta_\tau, \bz_\tau, \ba_\tau)$, where the attack actions $\ba_\tau$ are chosen by each method.
Note that $\tilde J(t)$ is computed on the actual instantiation of the environment data stream $\bz_0, \ldots, \bz_t$.
Better attack methods tend to have smaller $\tilde J(t)$.
We compare our algorithms with the following \textbf{Baseline Attackers}:

\textbf{Null Attack:} This is the baseline without attack, namely $\ba^{\textup{Null}}_t = \bz_t$ for all $t$.  We expect the null attack to form an upper bound on any attack method's empirical discounted cumulative cost $\tilde J(t)$.

\textbf{Greedy Attack:} The greedy strategy is applied widely as a practical heuristic in solving sequential decision problems (\cite{liu2017iterative,lessard2018optimal}). For our problem at time step $t$ the greedy attacker uses a time-invariant attack policy which minimizes the current step's running cost $g$. Specifically, the \textbf{greedy attack policy} can be written as
$
\ba^{\text{Greedy}}_t = \argmin_{\ba} g(\theta_{t},\bz_t,\ba).
$
If we instantiate $\gnef = \|\theta_t - \theta^\dagger\|^2_2$ for a target model $\theta^\dagger$ and $\gper = 0$, we exactly recover the algorithm in \cite{liu2017iterative}.
Both null attack and greedy attack can be viewed as time-invariant policies that do not utilize the information in $\hat P_t$.

\textbf{Clairvoyant Attack:} A clairvoyant attacker is an idealized attacker who knows the time horizon $T$
and the whole data sequence $\bz_{0:T-1}$ upfront.
In most realistic online data poisoning settings an attacker only know $\bz_{0:t}$ at time $t$.
Therefore, the clairvoyant attacker has strictly more information,
and we expect it to form a lower bound on realistic attack methods in terms of $\tilde J(t)$.
The clairvoyant attacker solves a finite time-horizon optimal control problem, equivalent to the formulation in \cite{wang2018data} but without terminal cost:
$\min_{\ba_{0:T-1}} \sum_{t=0}^{T-1} \gamma^{t} g(\theta_{t},\bz_t,\ba_t)$
subject to
$\theta_0$ given, $\bz_{0:T-1}$ given (clairvoyant), and
$\theta_{t+1} = f(\theta_t,\ba_t), t =  0 \ldots T-1$. 

\subsection{Poisoning Task Specification}
To specify a poisoning task is to define the victim learner $f$ in~\eqref{eq:f} and the attacker's running cost $g$ in~\eqref{eq:g_t}.
We evaluate all attacks on two types of victim learners: online logistic regression, a supervised learning algorithm, and online soft k-means clustering, an unsupervised learning algorithm.

\textbf{Online logistic regression:} Online logistic regression performs a binary classification task. The incoming data takes the form of $\bz_t = (\bx_t,y_t)$, where $\bx_t\in\R{^d}$ is the feature vector and $y_t\in\{-1,1\}$ is the binary label. In the experiments, we focus on attacking the feature part of the data, as is done in a number of prior works \cite{mei2015using,wang2018data,koh2017understanding}. The learner's update rule is one step of gradient descent on the log likelihood with step size $\eta$:
$
f\left(\theta,(\bx,y)\right) = \theta + \eta \frac{y\bx}{ 1+\exp(y\theta^\tp \bx)}.
$
The attacker wants to force the victim learner to stay close to a target parameter $\theta^{\dagger}$, i.e.\ this is a targeted attack.
The attacker's cost function $g$ is a weighted sum of two terms: 
the nefarious cost $\gnef$ is the negative cosine similarity between the victim's parameter and the target parameter,
and the perturbation cost $\gper$ is the $L_2$ distance between the perturbed feature vector and the clean one, i.e.\
$g\left(\theta_{t},(\bx_t,y),(\bx_t',y)\right) =  - \lambda \cos\left(\theta_t,\theta^{\dagger}\right) + \|\bx_t'-\bx_t\|^2$.
Recall $\cos(a,b) \defeq \frac{a^\tp b}{\|a\|\|b\|}$.

\textbf{Online soft k-means:} Online soft k-means performs a k-means clustering task. The incoming data contains only the feature vector, i.e.\ $\bz_t = \bx_t$. Its only difference from traditional k-means is that instead of updating only the centroid closest to the current data point, it updates all the centroids but the updates are weighted by their squared distances to the current data point using the softmax function \cite{bezdek1984fcm}. 
Specifically, the learner's update rule is one step of soft k-means update 
with step size $\eta$ on all centroids, i.e.\ $f(\theta^{(j)},\ba) = \theta^{(j)} + \eta  r_j (\ba-\theta^{(j)})$, $j=1,\dots,k$, where $\mathbf r = \mbox{softmax}\bigl(-\|\ba - \theta^{(1)}\|^2,\dots,-\|\ba - \theta^{(k)}\|^2\bigr)$.
Recall $\mbox{softmax}(x_1,\dots,x_k) \defeq \bigl[\frac{e^{x_1}}{\sum_j^k e^{x_j}},\dots,\frac{e^{x_k}}{\sum_j^k e^{x_j}}\bigr]^\tp$.
Similar to online logistic regression, we consider a targeted attack objective. The attacker wants to force the learned centroids to each stay close to the corresponding target centroid ${\theta^{\dagger}}^{(j)}$.
The attacker's cost function $g$ is a weighted sum of two terms: 
the nefarious cost function $\gnef$ is the sum of the squared distance between each of the victim's centroid and the corresponding target centroid,
and the perturbation cost $\gper$ is the $\ltwo$ distance between the perturbed feature vector and the clean one, i.e.\
$
g(\theta_t,\bz_t,\ba_t) =\lambda \sum^k_{j=1}\|f(\theta,\ba)^{(j)}-{\theta^{\dagger}}^{(j)}\|^2 +  \|\ba_t-\bz_t\|^2. 
$

\subsection{Synthetic Data Experiments}
We first show a synthetic data experiment where the attack policy can be visualized. 
The environment is a mixture of two 1D Gaussian: $P = \frac{1}{2}N(\theta^{(1)},1)+\frac{1}{2}N(\theta^{(2)},1)$  
with $\theta^{(1)} = -1$ and $\theta^{(2)} = +1$.
The victim learner is online soft k-means with $k=2$ and initial parameter $\theta_0^{(1)} = -2, \theta_0^{(2)} = +2$. The attack target is ${\theta^{\dagger}}^{(1)} = -3$ and ${\theta^{\dagger}}^{(2)} = +3$, namely the opposite of how the victim's parameters should move. 
We set the learning rate $\eta = 0.01$, cost regularizer $\lambda = 10$, discounting factor $\gamma = 0.99$, evaluation length $T=500$ and look-ahead horizon for MPC $h=100$. 
For attack methods that requires solving a nonlinear program, including GREEDY, NLP and Clairvoyant, we use the JuMP modeling language \cite{dunning2017jump} and the IPOPT
interior-point solver \cite{wachter2006implementation}.
Following the above specification, we run each attack method on the same data stream and compare their behavior. 
\begin{figure*}[t!]
	\centering
	\begin{minipage}[t]{0.25\columnwidth}
		\centering
		\includegraphics[width=\columnwidth]{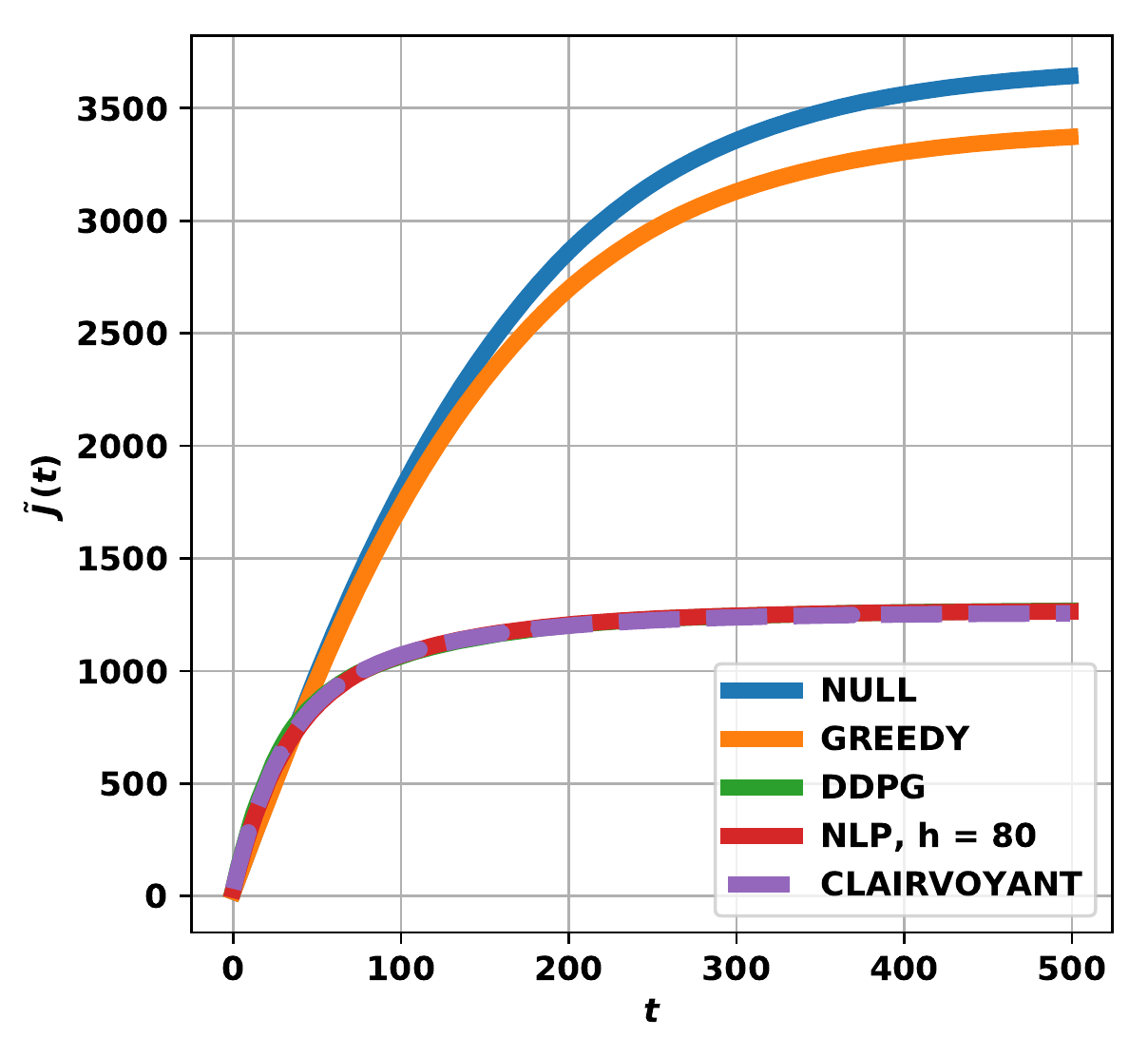} 
		\subcaption{Cumulative Attack Costs}
		\label{fig:toy_rewards}
	\end{minipage}%
	~
	\begin{minipage}[t]{0.25\columnwidth}
		\centering
		\includegraphics[width=.95\columnwidth]{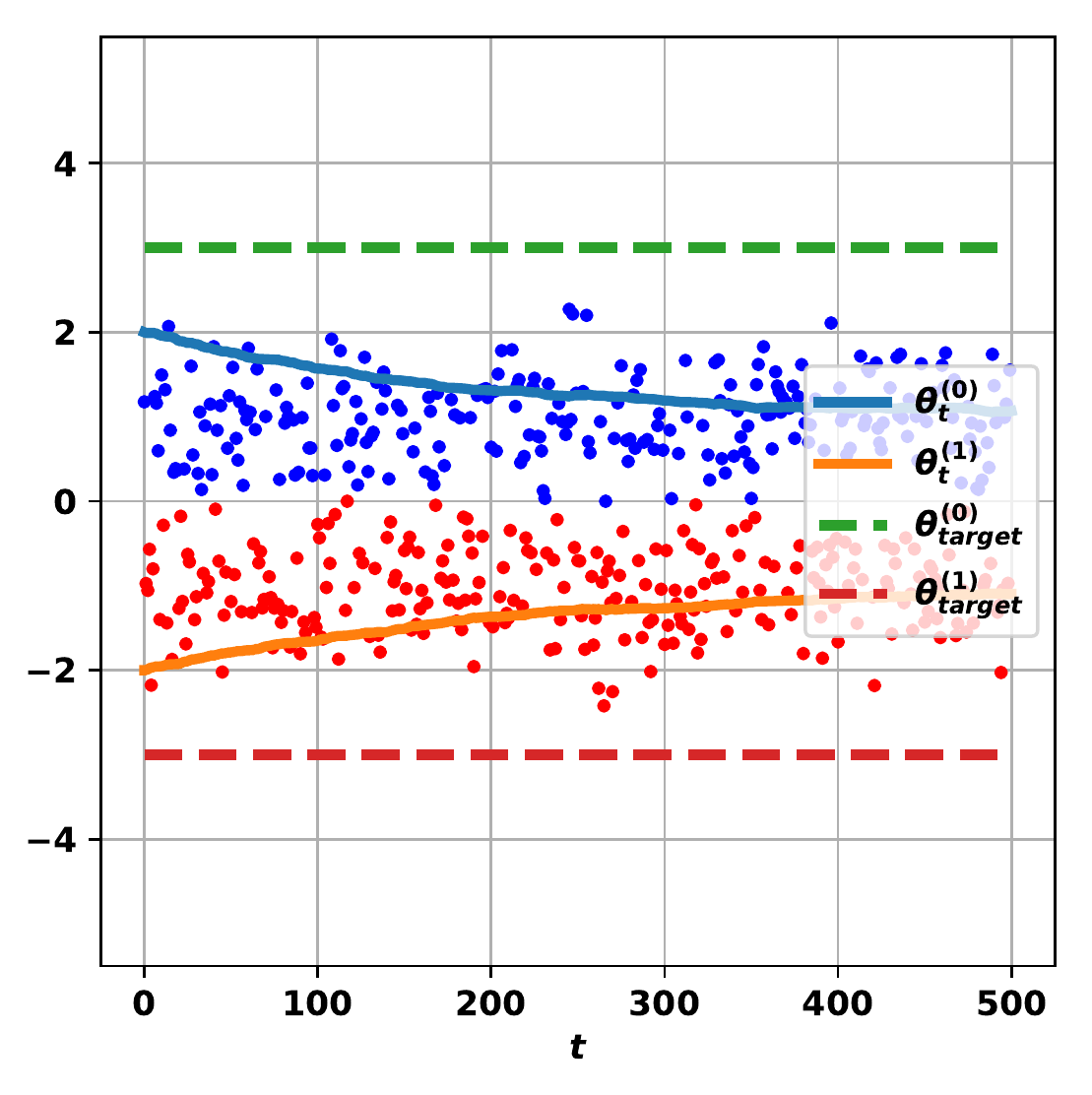} 
		\subcaption{NULL Attack}
		\label{fig:toy_null_attack}
	\end{minipage}%
	~
	\begin{minipage}[t]{0.25\columnwidth}
		\centering
		\includegraphics[width=.95\columnwidth]{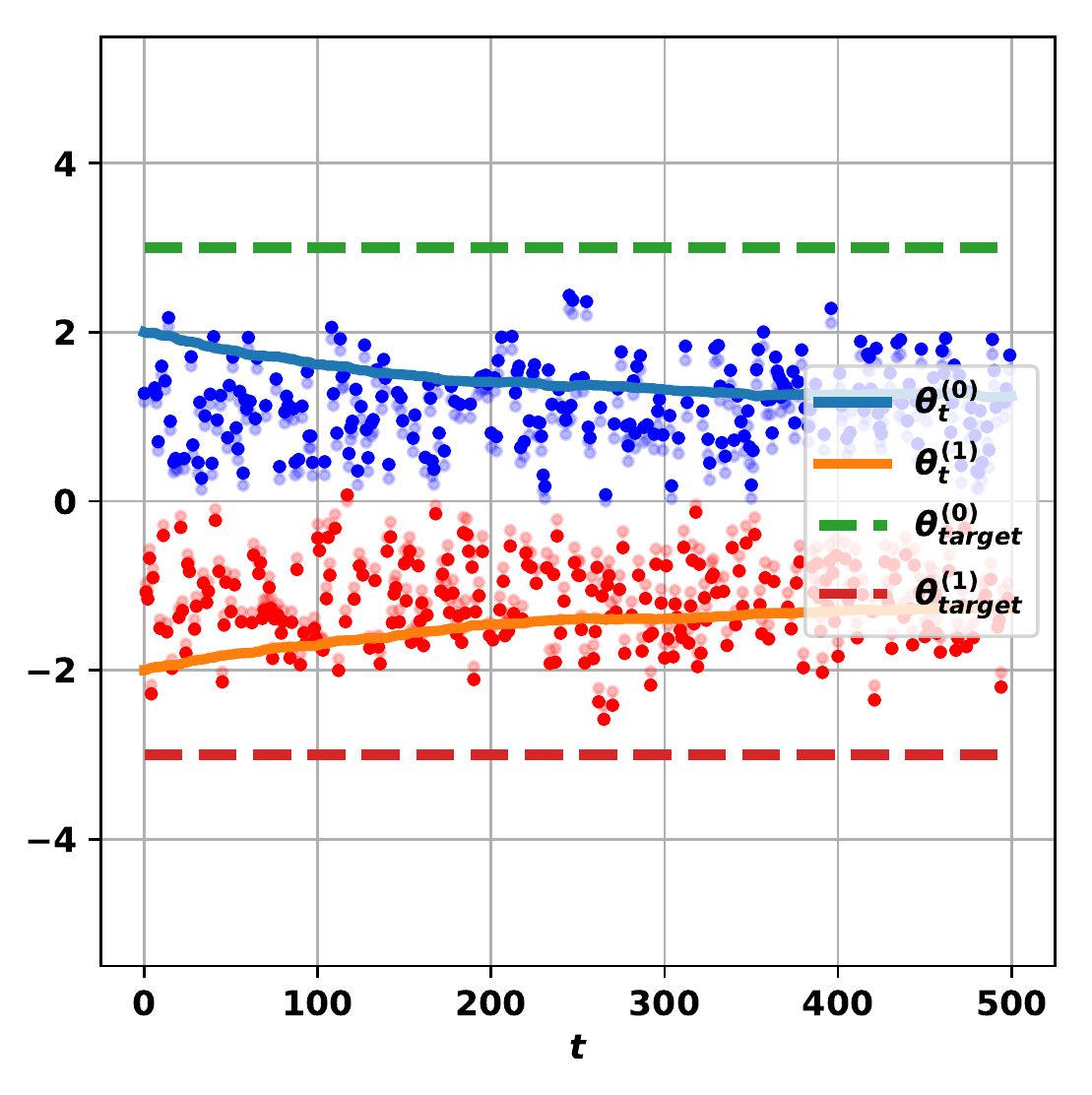} 
		\subcaption{GREEDY Attack}
		\label{fig:toy_greedy_attack}
	\end{minipage}%
~
\begin{minipage}[t]{0.25\columnwidth}
	\centering
	\includegraphics[width=.95\columnwidth]{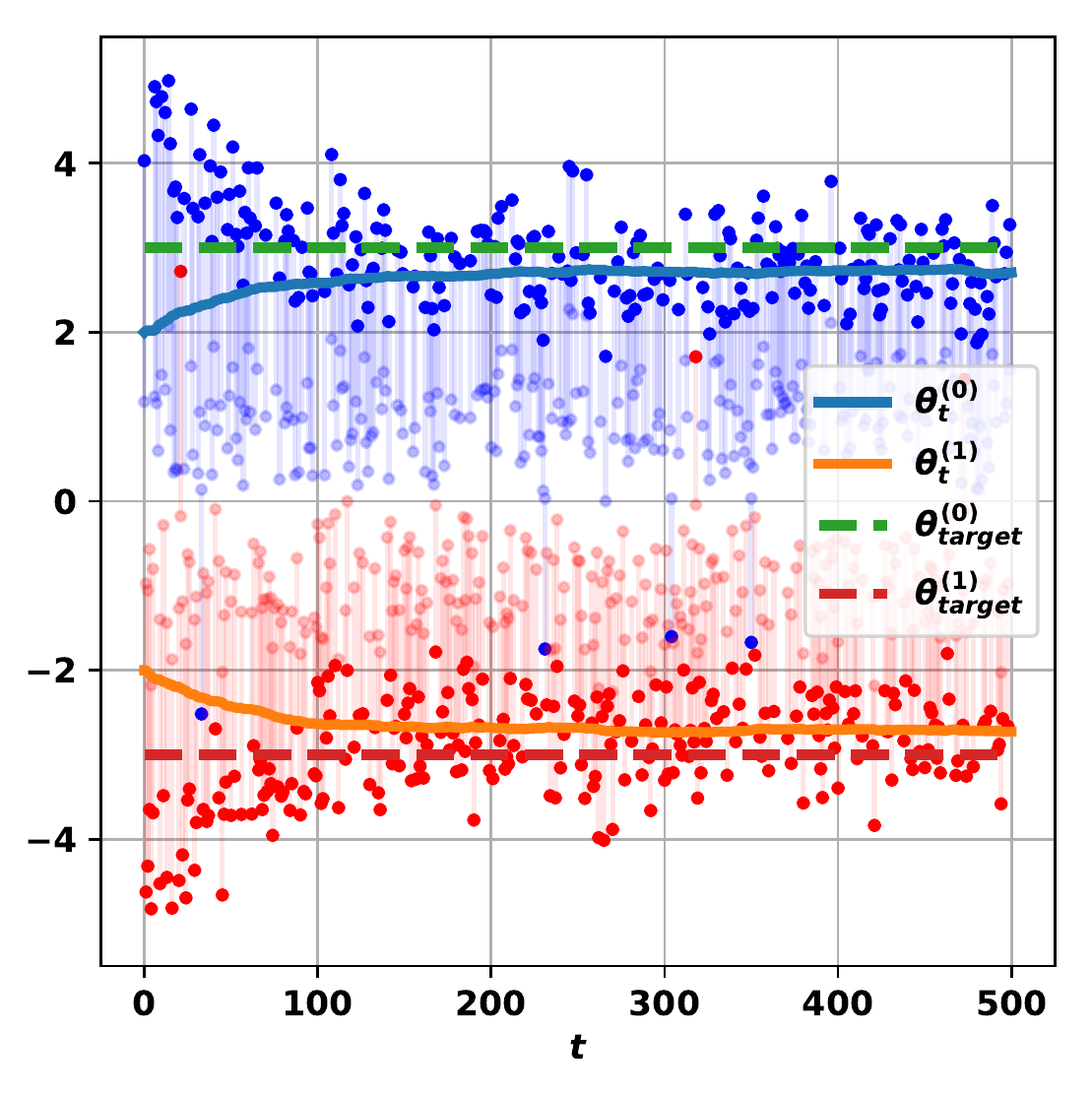} 
	\subcaption{NLP Attack}
	\label{fig:toy_MPC_attack}
\end{minipage}%
\caption{Synthetic data experiments. 
In (b)-(f), transparent blue and red dots indicate clean positive and negative data point $\bz_t$ at time step $t$, solid dots indicate attacker-perturbed data point $\ba_t$, vertical lines in between indicate the amount of perturbation.} 
	\label{fig:toy}
\end{figure*}

\textbf{Results: }
Figure~\ref{fig:toy_rewards} shows the empirical discounted cumulative cost $\tilde J(t)$ as the attacks go on. On this toy example, the null attack baseline achieves $\tilde J(T)=3643$ at $T=500$. The greedy attacker is only slight more effective at $\tilde J(T)=3372$. 
NLP and DDPG (curve largely overlap and hidden under NLP)   achieve $1265$ and $1267$, respectively, almost matching Clairvoyant's $1256$.
As expected, the null and clairvoyant attacks form upper and lower bounds on $\tilde J(t)$.

Figure~\ref{fig:toy}b-f shows the victim's $\theta_t$ trajectory as attacks go on. 
Without attack (null), $\theta_t$ converges to the true parameter $-1$ and $+1$.
The greedy attack only perturbs each data point slightly, failing to force $\theta_t$ toward attack targets.
This failure is due to its greedy nature: the immediate cost $g_t$ at each round is indeed minimized, but not enough to move the model parameters close to the target parameters.
In contrast, NLP and DDPG (trajectory similar to NLP, not shown) exhibit a different strategy in the earlier rounds. They inject larger perturbations to the data points and sacrifice larger immediate costs in order to drive the victim's model parameters quickly towards the target. In later rounds they only need to stabilize the victim's parameters near the target with smaller per-step cost. 


\subsection{Real Data Experiments}

In the real data experiments, we run each attack method on 10 data sets across two victim learners. 

\textbf{Datasets: } We use 5 datasets for online logistic regression: Banknote Authentication (with feature dimension $d=4$), Breast Cancer ($d=9$), Cardiotocography ($d=25$), Sonar ($d=60$), and MNIST 1 vs.\ 7 ($d = 784$), and 5 datasets for online k-means clustering: User Knowledge ($d=6, k=2$), Breast Cancer ($d=10, k=2$), Seeds ($d = 8,k=3$), posture ($d=11,k=5$), MNIST 1 vs.\ 7 ($d = 784, k=2$). All datasets except for MNIST can be found in the UCI Machine Learning Repository \cite{Dua:2019}. Note that two datasets, Breast Cancer and MNIST, are shared across both tasks.

\textbf{Preprocessing: }
To reduce the running time, for datasets with dimensionality $d>30$, we reduce the dimension to $30$ via PCA projection. 
Then, all datasets are normalized so that each feature has mean 0 and variance 1. Each dataset is then turned into a data stream by random sampling. Specifically, each training data point $\bz_t$ is sampled uniformly from the dataset with replacement.

\textbf{Experiment Setup: }
In order to demonstrate the general applicability of our methods, we draw both the victim's initial model $\theta_0$ and the attacker's target $\theta^{\dagger}$ at random from a standard Gaussian distribution of the appropriate dimension, for both online logistic regression and online k-means in all 10 datasets. 
Across all datasets, we use the following hyperparameters: $\eta=0.01,\gamma=0.99,T=300$. For online logistic regression $\lambda = 100$ while for online k-means $\lambda=10$. 

For DDPG attacker we only perform policy learning at the beginning to obtain $\phi_{\hat{\M_0}}$; the learned policy is then fixed and used to perform all the attack actions in later rounds. In order to give it a fair chance, we give it a pre-attack dataset $\bz_{-n:-1}$ of size $n=1000$. For the sake of fair comparisons, we give the same pre-attack dataset to NLP as well. 
For NLP attack we set the look-ahead horizon $h$ such that the total runtime to perform $T=300$ attacks does not exceed the DDPG training time, which is 24 hours on an Intel Core i7-6800K CPU \@ 3.40GHz with 12 cores. 
This results in $h = 20$ for online logistic regression on CTG, Sonar and MNIST, and $h=80$ in all other experiments.

\begin{figure*}[t!]
	\centering
	\begin{minipage}[t]{0.20\columnwidth}
		\centering
		\includegraphics[width=\columnwidth]{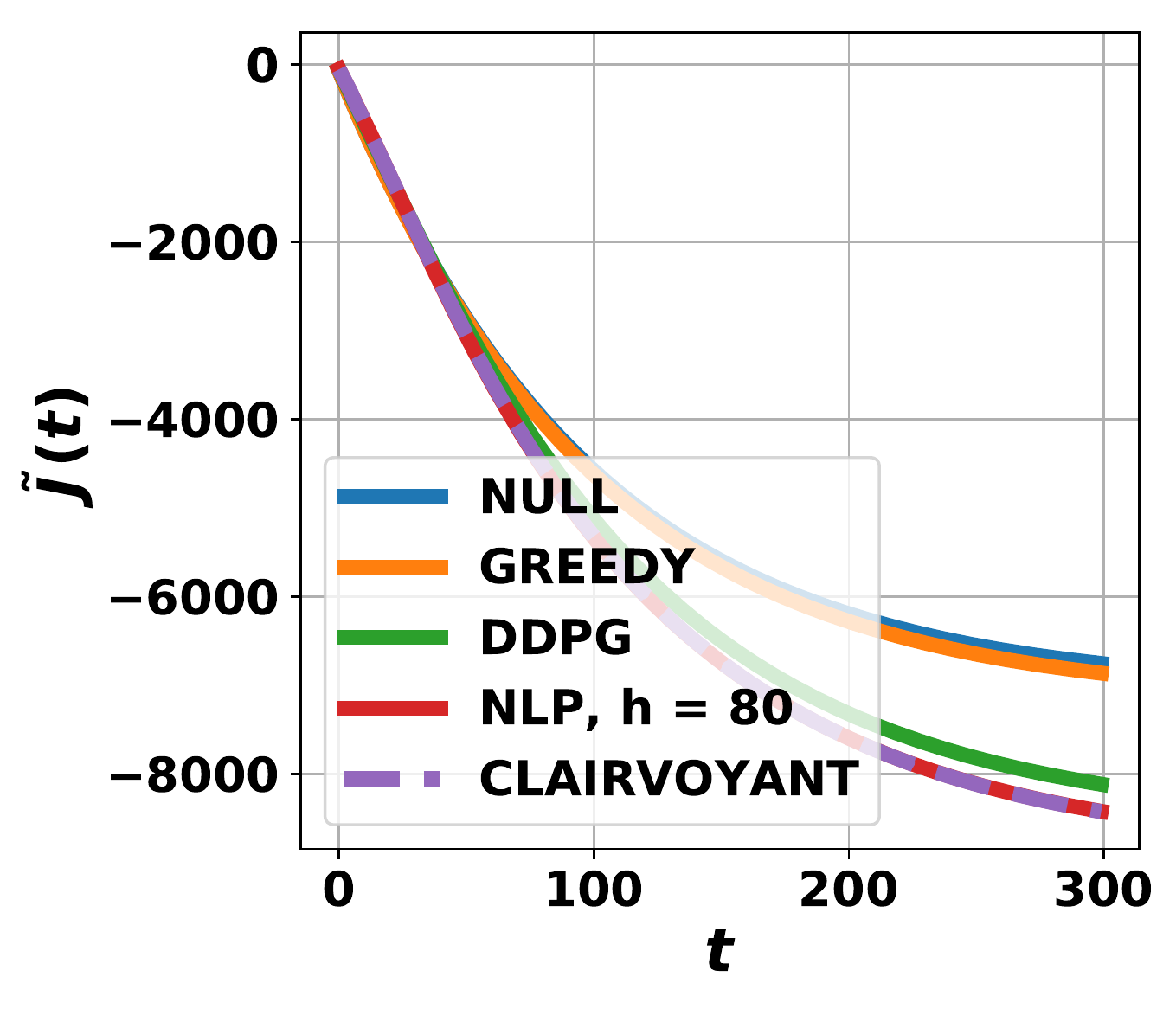} 
		\subcaption{Banknote}
		\label{fig:banknote_lr}
	\end{minipage}%
	\begin{minipage}[t]{0.20\columnwidth}
		\centering
		\includegraphics[width=\columnwidth]{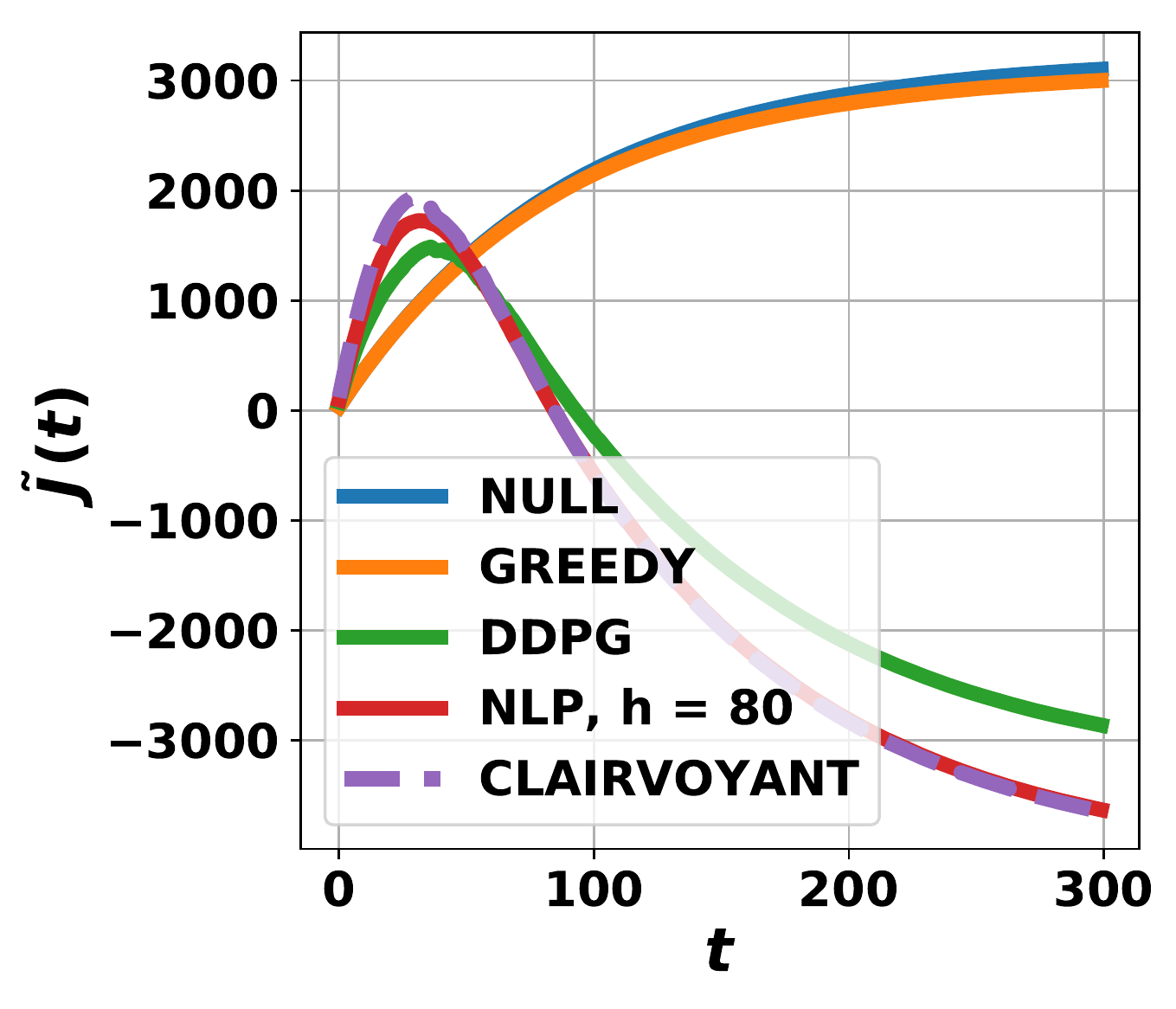} 
		\subcaption{Breast}
		\label{fig:breast_lr}
	\end{minipage}%
	\begin{minipage}[t]{0.20\columnwidth}
		\centering
		\includegraphics[width=\columnwidth]{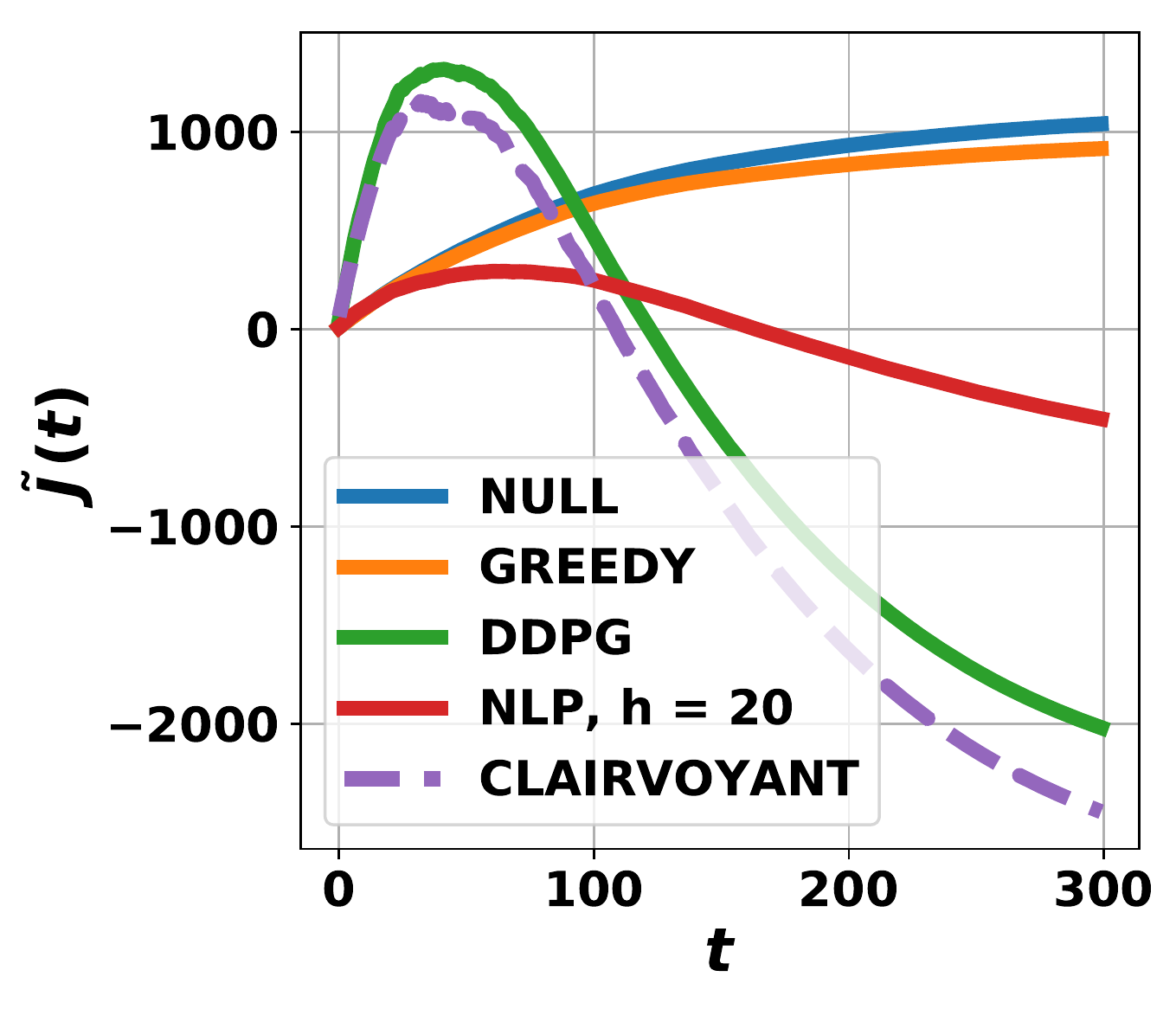} 
		\subcaption{CTG}
		\label{fig:CTG_lr}
	\end{minipage}%
	\begin{minipage}[t]{0.20\columnwidth}
		\centering
		\includegraphics[width=\columnwidth]{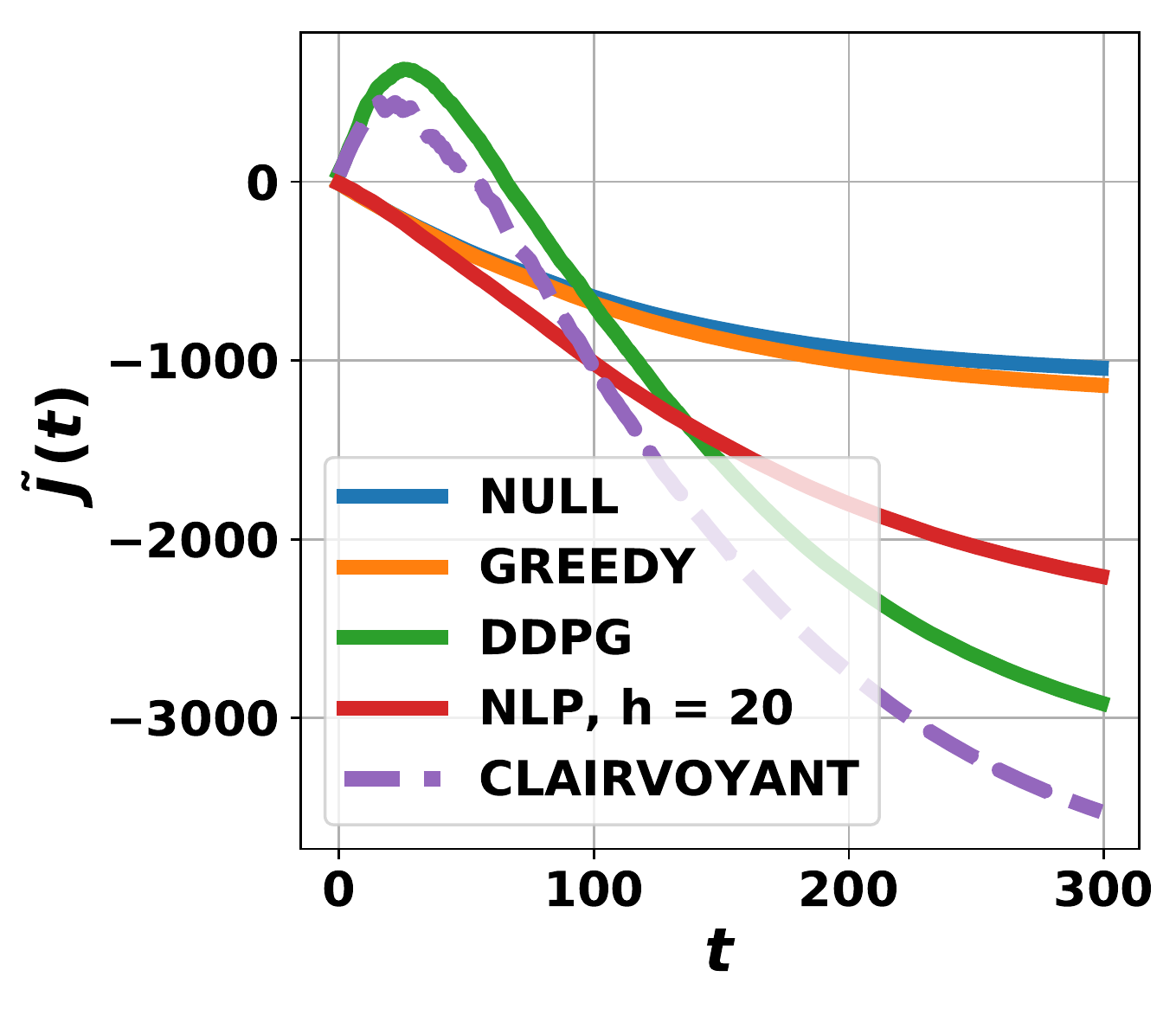} 
		\subcaption{Sonar}
		\label{fig:sonar_lr}
	\end{minipage}%
	\begin{minipage}[t]{0.20\columnwidth}
		\centering
		\includegraphics[width=\columnwidth]{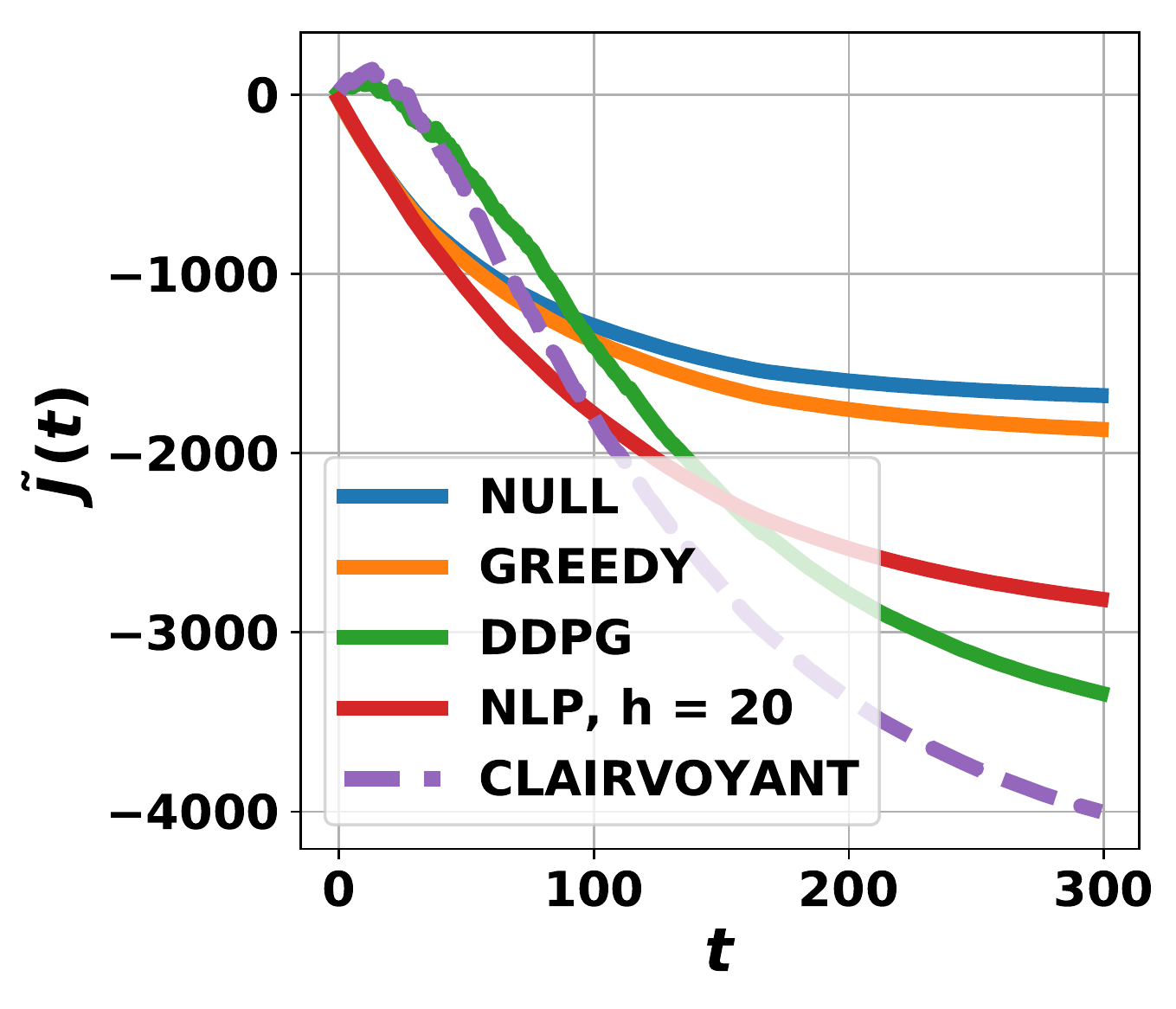} 
		\subcaption{MNIST 1 vs.\ 7}
		\label{fig:mnist_lr}
	\end{minipage}%
	\\
	\begin{minipage}[t]{0.20\columnwidth}
		\centering
		\includegraphics[width=\columnwidth]{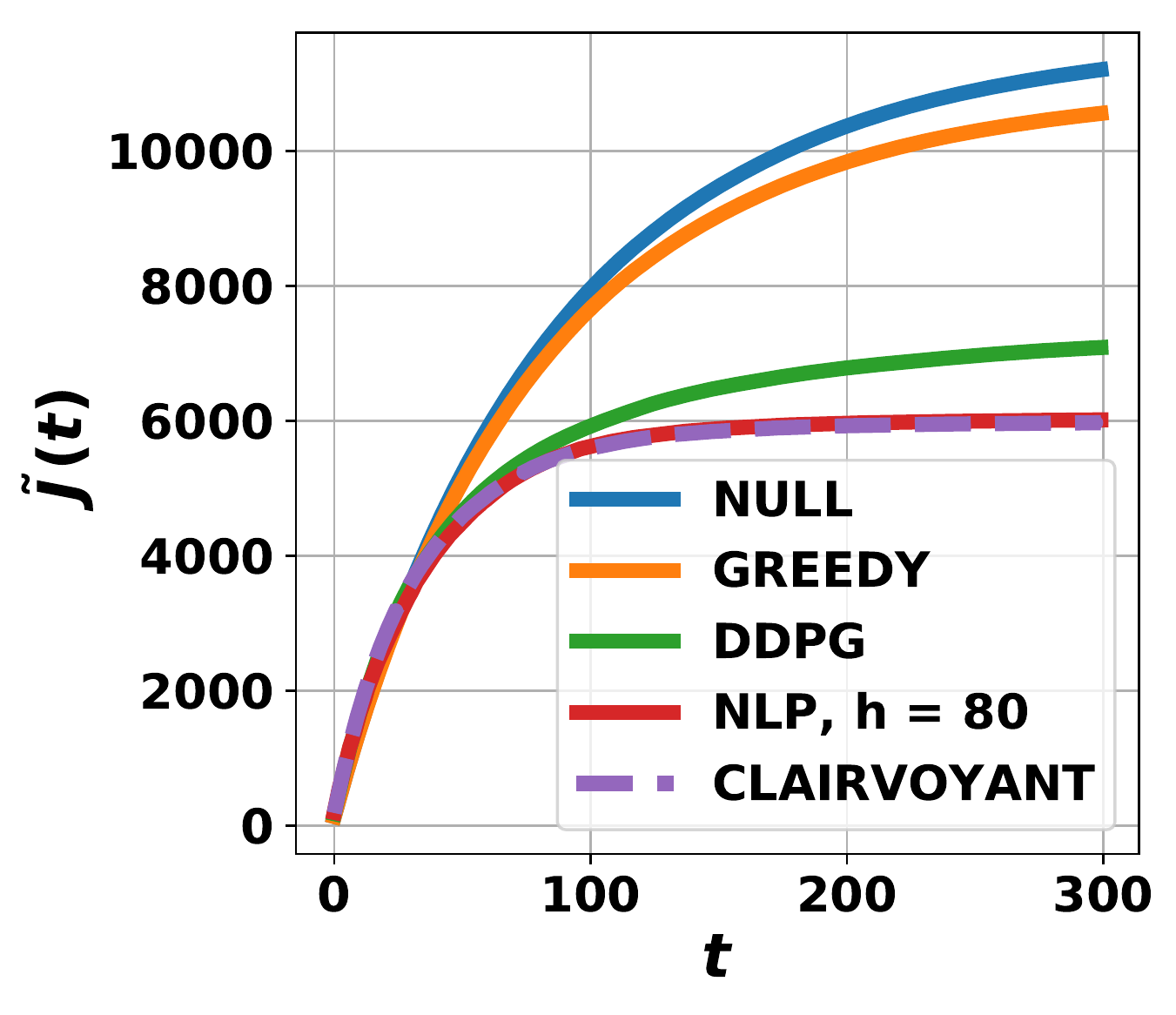} 
		\subcaption{Knowledge}
		\label{fig:knowledge_km}
	\end{minipage}%
	\begin{minipage}[t]{0.20\columnwidth}
		\centering
		\includegraphics[width=\columnwidth]{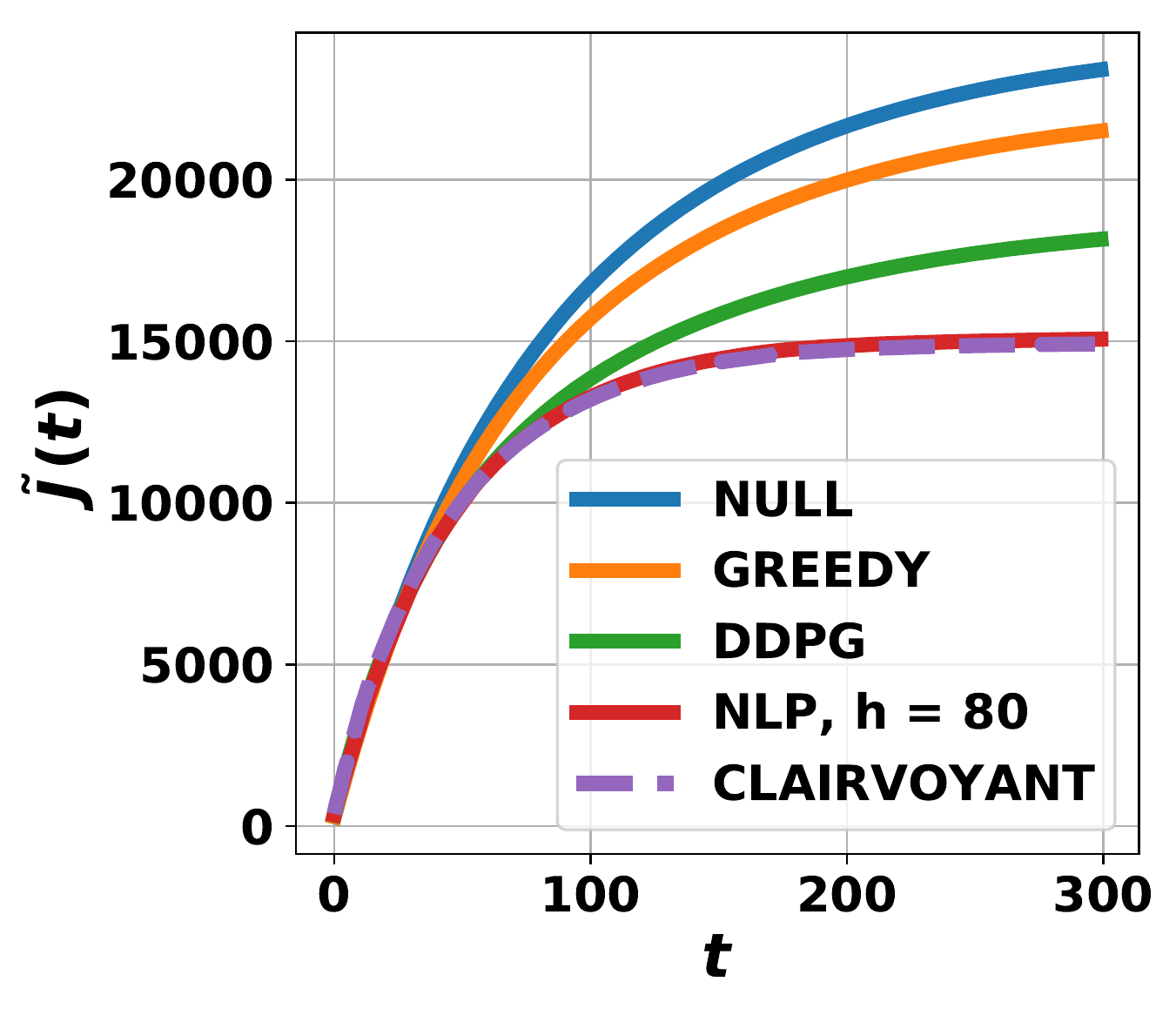} 
		\subcaption{Breast}
		\label{fig:breast_km}
	\end{minipage}%
	\begin{minipage}[t]{0.20\columnwidth}
		\centering
		\includegraphics[width=\columnwidth]{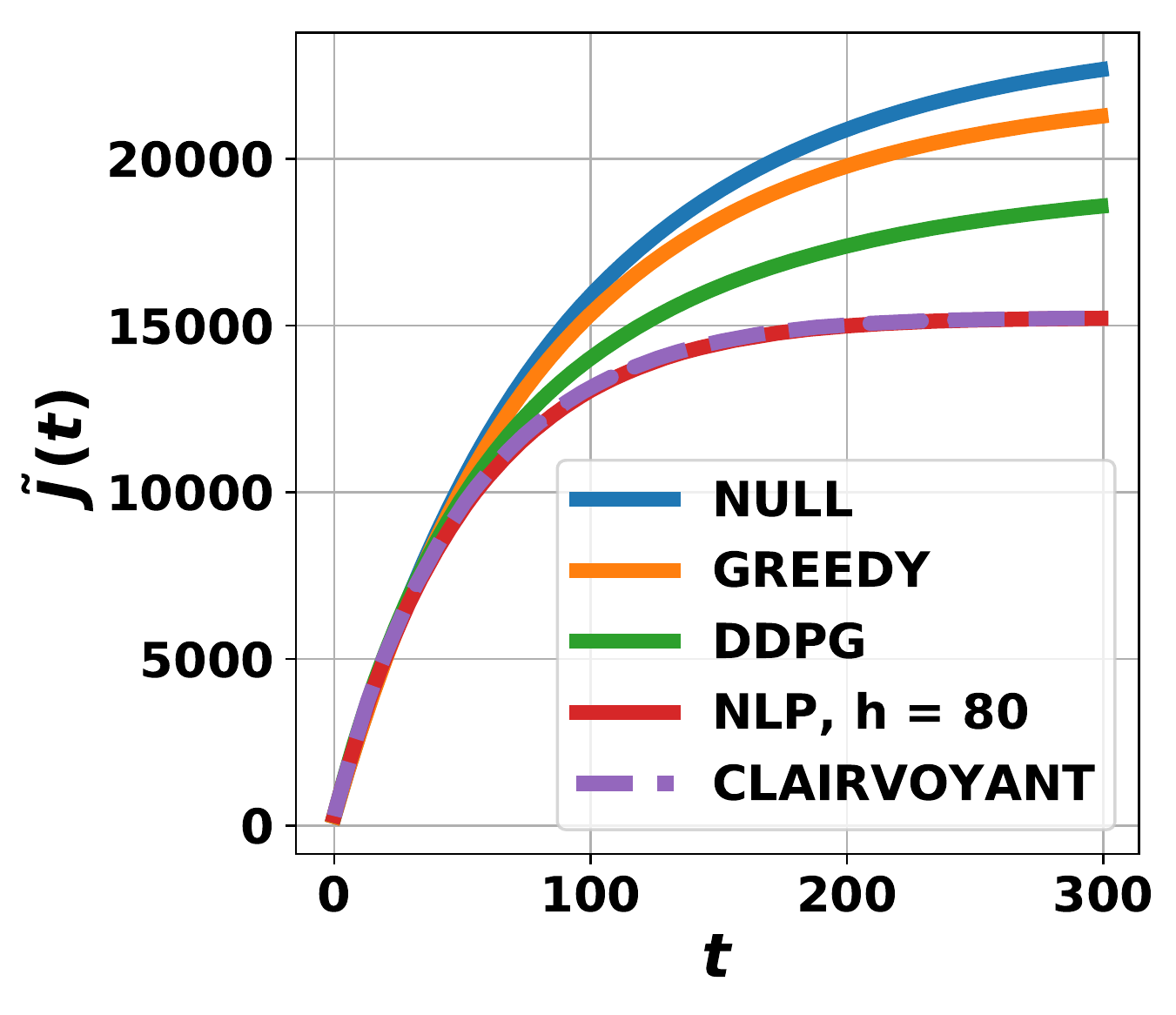} 
		\subcaption{Seeds}
		\label{fig:seeds_km}
	\end{minipage}%
	\begin{minipage}[t]{0.20\columnwidth}
		\centering
		\includegraphics[width=\columnwidth]{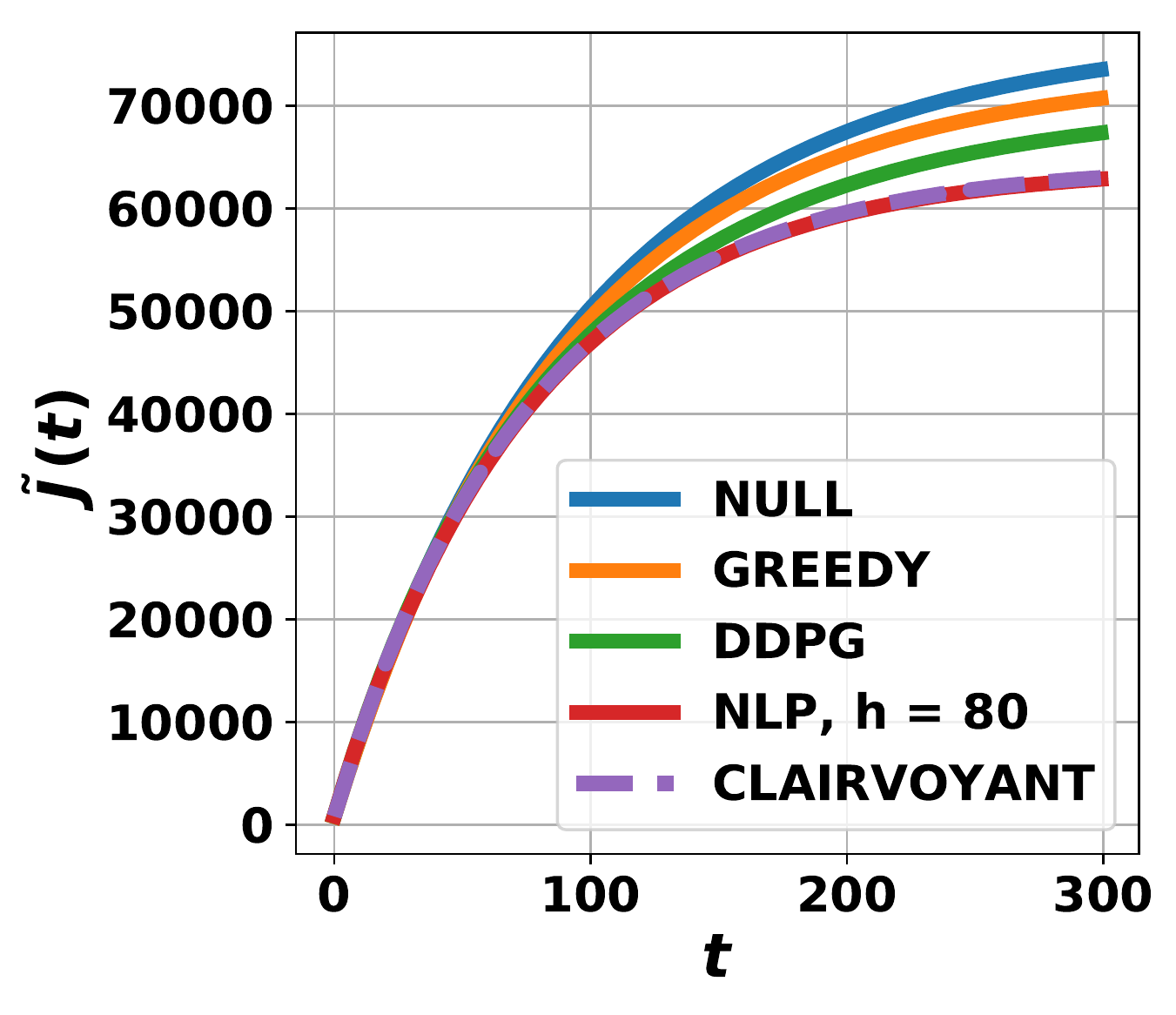} 
		\subcaption{Posture}
		\label{fig:posture_km}
	\end{minipage}%
	\begin{minipage}[t]{0.20\columnwidth}
		\centering
		\includegraphics[width=\columnwidth]{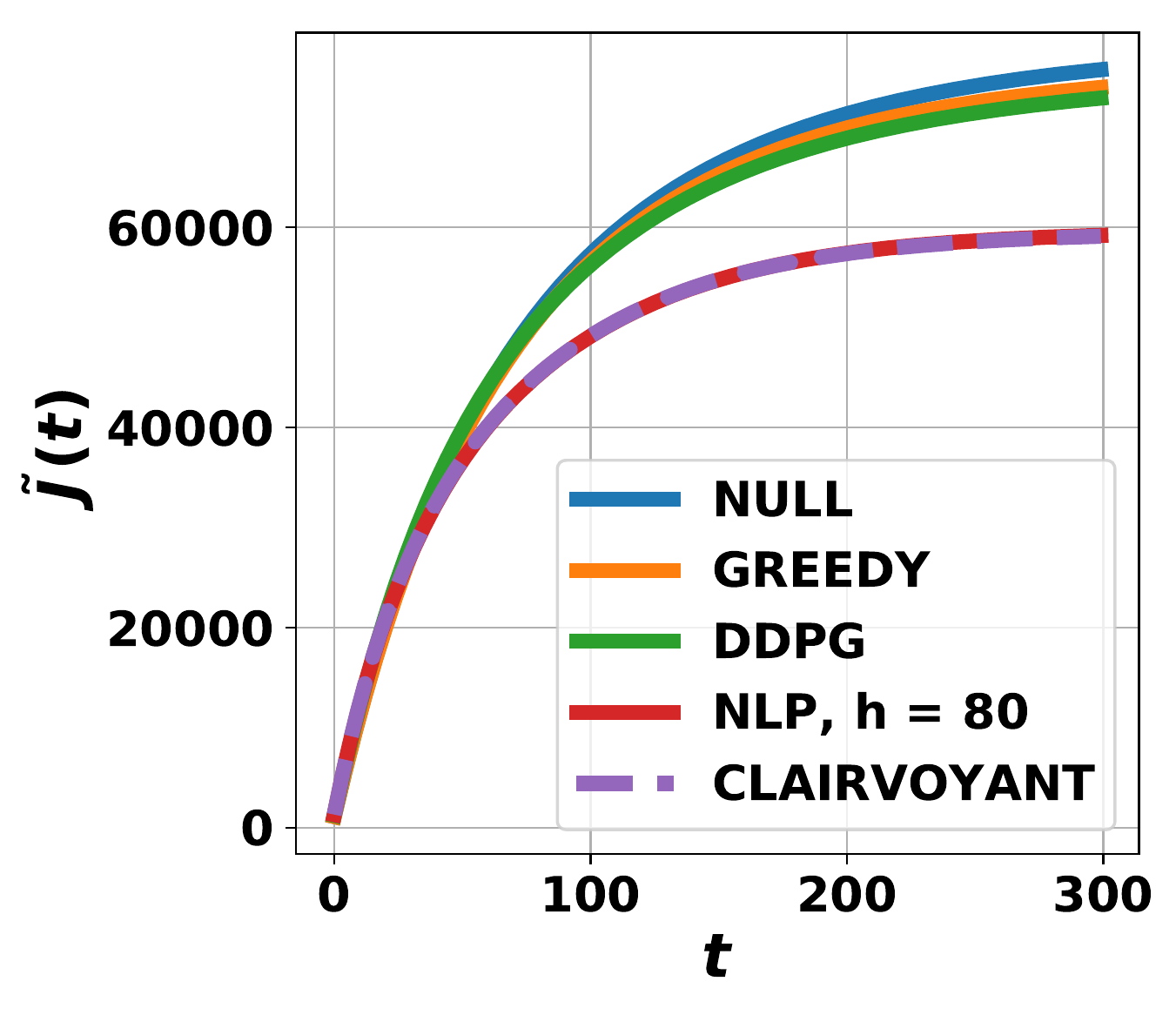} 
		\subcaption{MNIST 1 vs.\ 7}
		\label{fig:mnist_km}
	\end{minipage}%
	\caption{The empirical discounted cumulative reward $\tilde J(t)$ for the five attack methods across 10 real datasets. The first row is on online logistic regression and the second row is on online k-means. Note that $g(\cdot)$ for online logistic regression can be negative, and thus the $\tilde J(t)$ curve can be decreasing.}
	\label{fig:real}
\end{figure*}

\textbf{Results: } The experiment results are shown in figure \ref{fig:real}. Interestingly, several consistent patterns emerge from the experiments: The clairvoyant attacker consistently achieves the lowest cumulative cost $\tilde J(T)$ across all 10 datasets. This is not surprising, as the clairvoyant attacker has extra information of the future. The NLP attack achieves clairvoyant-matching performance on all 7 datasets in which it is given a large enough look-ahead horizon, i.e.\ $h=80$. 
DDPG follows closely next to MPC and Clairvoyant on most of the datasets, indicating that the pre-trained policy $\phi_{\hat{\M_0}}$ can achieve reasonable attack performance in most cases. On the 3 datasets where $h=20$ for NLP, DDPG exceeds the short-sighted NLP, indicating that when the computational resource is limiting, DDPG has an advantage by avoiding the iterative retraining that NLP cannot bypass. GREEDY does not do well on any of the 10 datasets, achieving only a slightly lower cost than the NULL baseline. This matches our observations in the synthetic experiment. 

Each of the attack methods also exhibits strategic behavioral patterns similar to what we observe in the synthetic experiment. In particular, the optimal-control based methods NLP and DDPG sacrifice larger immediate costs in earlier rounds in order to achieve smaller attack costs in later rounds. 
This is especially obvious in the online logistic regression plots \ref{fig:real}b-e, where the cumulative costs $\tilde{J}(t)$ rise dramatically in the first 50 rounds, becoming higher than the cost of NULL and GREEDY around that time. This early sacrifice pays off after $t=50$ where the cumulative cost starts to fall much faster. 
In \ref{fig:real}c-e, however, the short-sighted NLP (with $h=20$) fails to fully pick up this long-term strategy, and exhibits a behavior close to an interpolation of greedy and optimal. This is not surprising, as NLP with horizon $h=1$ is indeed equivalent to the GREEDY method. Thus, there is a spectrum of methods between GREEDY and NLP that can achieve various levels of performance with different computational costs.

\section{Conclusion}\label{sec:disc}
In this paper, we formulated online poisoning attacks as a stochastic optimal control problem. 
We proposed two attack algorithms: a model-based planning approach and a model-free reinforcement learning approach, and showed that both are able to achieve near clairvoyant-levels of performance. 
We also provided analysis to characterize the optimality gap between a realistic attacker with no knowledge of $P$ and a clairvoyant attacker that knows $P$ in advance. 

\subsubsection*{Acknowledgments}
This work is supported in part by NSF 1750162, 1837132, 1545481, 1704117, 1623605, 1561512, the MADLab AF Center of Excellence FA9550-18-1-0166, and the University of Wisconsin.

\bibliography{neurips_2019}
\bibliographystyle{plain}

\newpage

\section{Appendix}
These proofs follow the technique in Nan Jiang's Statistical Reinforcement Learning lecture notes (\url{https://nanjiang.cs.illinois.edu/cs598/}).
\subsection{Proof of theorem \ref{thm:sim_lemma}}
\begin{proof}
	For any policy $\phi$ and state $s\in S$, we have
	\begin{eqnarray}
	&&|V_\mathcal{\hat M}^\phi(s) - V_\mathcal{M}^\phi(s)|\\
	&\stackrel{\text{Bellman}}{=}& |g(s,\phi(s)) + \gamma \E_{\hat T( s' | s, \phi(s))} V_\mathcal{\hat M}^\phi(s') - g(s,\phi(s)) - \gamma \E_{T( s' | s, \phi(s))} V_\mathcal{M}^\phi(s')|\nonumber\\
	&=& \gamma|\E_{\hat T( s' | s, \phi(s))} V_\mathcal{\hat M}^\phi(s') - 
	\E_{T( s' | s, \phi(s))} V_\mathcal{M}^\phi(s')|\nonumber\\
	&=& \gamma|\E_{\hat T( s' | s, \phi(s))} V_\mathcal{\hat M}^\phi(s') - 
	\E_{T( s' | s, \phi(s))} V_\mathcal{\hat M}^\phi(s') + 
	\E_{T( s' | s, \phi(s))} V_\mathcal{\hat M}^\phi(s') -
	\E_{T( s' | s, \phi(s))} V_\mathcal{M}^\phi(s')|\nonumber\\
	&\stackrel{\text{tri.}}{\leq}& \gamma |\E_{\hat T( s' | s, \phi(s))} V_\mathcal{\hat M}^\phi(s') - 
	\E_{T( s' | s, \phi(s))} V_\mathcal{\hat M}^\phi(s')| + \gamma|\E_{T( s' | s, \phi(s))} V_\mathcal{\hat M}^\phi(s') -
	\E_{T( s' | s, \phi(s))} V_\mathcal{M}^\phi(s')|\nonumber\\
	&\stackrel{\text{extremal}}{\leq}& \gamma |\E_{\hat T( s' | s, \phi(s))} V_\mathcal{\hat M}^\phi(s') - \E_{T( s' | s, \phi(s))} V_\mathcal{\hat M}^\phi(s')| 
	+ \gamma \sup_{s\in S}|V_\mathcal{\hat M}^\phi(s) - V_\mathcal{M}^\phi(s)| \nonumber\\
	&{=}& \gamma \left|\left\langle \hat T( \cdot | s, \phi(s)) - T( \cdot | s, \phi(s)), V_\mathcal{\hat M}^\phi(\cdot) \right\rangle\right| 
	+ \gamma \sup_{s\in S}|V_\mathcal{\hat M}^\phi(s) - V_\mathcal{M}^\phi(s)| \nonumber\\
	&\stackrel{\text{const.}}{=}& \gamma \left|\left\langle \hat T( \cdot | s, \phi(s)) - T( \cdot | s, \phi(s)), V_\mathcal{\hat M}^\phi(\cdot) - \frac{C_{\max}}{2(1-\gamma)}\right\rangle\right| 
	+ \gamma \sup_{s\in S}|V_\mathcal{\hat M}^\phi(s) - V_\mathcal{M}^\phi(s)| \nonumber\\
	&\stackrel{\mbox{\scriptsize{H\"older}}}{\leq}& \gamma \|\hat T(\cdot | s, \phi(s)) - T(\cdot | s,\phi(s))\|_1 \sup_{s\in S}\left|V_\mathcal{\hat M}^\phi(s) - \frac{C_{\max}}{2(1-\gamma)}\right| + \gamma \sup_{s\in S}|V_\mathcal{\hat M}^\phi(s) - V_\mathcal{M}^\phi(s)|\nonumber\\
	&\stackrel{\text{range}}{\le}& \gamma \|\hat T(\cdot | s, \phi(s)) - T(\cdot | s,\phi(s))\|_1 \frac{C_{\max}}{2(1-\gamma)} + \gamma \sup_{s\in S}|V_\mathcal{\hat M}^\phi(s) - V_\mathcal{M}^\phi(s)|\nonumber\\
	&=& \gamma \|\hat P - P\|_1 \frac{C_{\max}}{2(1-\gamma)} + \gamma \sup_{s\in S}|V_\mathcal{\hat M}^\phi(s) - V_\mathcal{M}^\phi(s)|\nonumber\\
	&\leq& \frac{\gamma C_{\max} \epsilon}{2(1-\gamma)} + \gamma \sup_{s\in S}|V_\mathcal{\hat M}^\phi(s) - V_\mathcal{M}^\phi(s)|. \nonumber
	\end{eqnarray}
	Since this holds for all $s \in S$, we can also take the supremum on the LHS, which yields 
	\begin{eqnarray}
	\sup_{s\in S} |V_\mathcal{\hat M}^\phi(s) - V_\mathcal{M}^\phi(s)|\leq \frac{\gamma C_{\max}\epsilon }{2(1-\gamma)^2}.\label{eq:sim_lemma}
	\end{eqnarray}
	Now, for any $s\in S$,
	\begin{eqnarray}
	V_\M^{\phi^\star_{\hat{\M}}}(s) - V_\M^{\phi^\star_\M}(s) &=& 
	V_\M^{\phi^\star_{\hat{\M}}}(s) - V_{\hat\M}^{\phi^\star_{\M}}(s) + V_{\hat\M}^{\phi^\star_\M}(s) - V_\M^{\phi^\star_\M}(s)\\
	&\stackrel{\text{opt.}}{\leq}& V_{\M}^{\phi^\star_{\hat\M}}(s) - \left(V_{\hat\M}^{\phi^\star_{\hat\M}}(s) \right) + V_{\hat\M}^{\phi^\star_{\M}}(s) - V_\M^{\phi^\star_{\M}}(s)\\
	&\stackrel{\eqref{eq:sim_lemma}}{\leq}& \frac{\gamma C_{\max}\epsilon }{2(1-\gamma)^2} + \frac{\gamma C_{\max}\epsilon }{2(1-\gamma)^2} \\
	&=& \frac{\gamma C_{\max}\epsilon }{(1-\gamma)^2}.
	\end{eqnarray}
	This completes the proof.
\end{proof}

\subsection{Proof of theorem \ref{thm:mn_convergence}}
\begin{proof}
	We first want to establish an $\ell_1$ concentration bound for multinomial distribution. Observe that for any vector $v\in \R^N$,
	\begin{eqnarray}
	\|v\|_1 = \max_{u\in\{-1,1\}^N} u^\tp v.
	\end{eqnarray}
	The plan is to prove concentration for each $u^\tp v$ first, and then union bound over all $u$ to obtain the $\ell_1$ error bound. Observe that $u^\tp \hat P$ is the average of $n$ i.i.d.\ random variables $u^\tp e_{x_i}$ with range $[-1,1]$. Then, by Hoeffding's Inequality, with probability at least $1-\delta/2^N$, we have
	\begin{eqnarray}
	u^\tp(\hat P-P)\leq 2\sqrt{\frac{1}{2n}\ln{2^{N+1}\over \delta}}.
	\end{eqnarray}
	Then, we can apply union bound across all $u\in\{-1,1\}^N$and get that, with probability at least $1-\delta$,
	\begin{eqnarray}
	\|\hat P - P\|_1 = \max_u u^\tp (\hat P - P) \leq 2\sqrt{\frac{1}{2n}\ln{2^{N+1}\over \delta}}
	\end{eqnarray}
	Substituting this quantity into Lemma~\ref{thm:sim_lemma} yields the desired result.
\end{proof}

\end{document}